\documentclass[runningheads]{llncs}
\usepackage{graphicx}
\usepackage{array}
\newcolumntype{C}[1]{>{\centering\let\newline\\\arraybackslash\hspace{0pt}}m{#1}}
\usepackage{amsmath}
\usepackage{amssymb}
\usepackage{gensymb}
\usepackage{booktabs}
\begin{document}
\title{GAttANet: Global attention agreement for convolutional neural networks\thanks{Supported by ANITI ANR grant ANR-19-PI3A-0004, AI-REPS ANR grant ANR-18-CE37-0007-01 and OSCI-DEEP ANR grant ANR-19-NEUC-0004.}}
\titlerunning{Global attention agreement networks}
%
\author{Rufin VanRullen\inst{1,2} \and Andrea Alamia\inst{1}}
\authorrunning{R. VanRullen and A. Alamia}
%
\institute{CerCo, CNRS, 31052 Toulouse, France \email{\{rufin.vanrullen,andrea.alamia\}@cnrs.fr} \and
ANITI, Universit\'e de Toulouse, 31062 Toulouse, France\\
}
\maketitle              
\vspace{-8mm}
\begin{abstract} 
Transformer attention architectures, similar to those developed for natural language processing, have recently proved efficient also in vision, either in conjunction with or as a replacement for convolutional layers. Typically, visual attention is inserted in the network architecture as a (series of) feedforward self-attention module(s), with mutual key-query agreement as the main selection and routing operation. However efficient, this strategy is only vaguely compatible with the way that attention is implemented in biological brains: as a separate and unified network of attentional selection regions, receiving inputs from and exerting modulatory influence on the entire hierarchy of visual regions. Here, we report experiments with a simple such attention system that can improve the performance of standard convolutional networks, with relatively few additional parameters. Each spatial position in each layer of the network produces a key-query vector pair; all queries are then pooled into a global attention query. On the next iteration, the match between each key and the global attention query modulates the network's activations---emphasizing or silencing the locations that agree or disagree (respectively) with the global attention system. We demonstrate the usefulness of this brain-inspired Global Attention Agreement network (GAttANet) for various convolutional backbones (from a simple 5-layer toy model to a standard ResNet50 architecture) and datasets (CIFAR10, CIFAR100, Imagenet-1k). Each time, our global attention system improves accuracy over the corresponding baseline.

\end{abstract}
\vspace{-8mm}
\section{Introduction}
\subsubsection{Transformer attention networks -}
Modern Natural Language Processing (NLP) strongly relies on attention mechanisms to handle long-distance relations between elements in a sequence of text. In particular, the Transformer architecture, which uses key-query vector agreement to determine information routing in feed-forward attention modules, has become an important component of most state-of-the-art language models~\cite{vaswani2017}.

More recently, the same Transformer attention strategy has been successfully integrated in state-of-the-art vision architectures. Some studies have proposed to insert separate attention modules within standard convolutional backbones~\cite{wang2017,bello2019}, while others have suggested to do away with convolutions entirely, and instead rely solely on Transformer operations~\cite{bello2019,ramachandran2019,zhao2020}. Indeed, it can be demonstrated that convolutions are actually a subset of all operations permitted by Transformer modules~\cite{cordonnier2020relationship}---so attention is strictly more expressive than convolution, although it may be less computationally efficient, depending on implementation. 

On the one hand, the latest vision Transformer architectures often surpass the performance of convolutional networks on image classification~\cite{dosovitskiy2020image,touvron2020training}. On the other hand, performance need not be the \emph{only} standard by which we should evaluate vision models. For instance, biological plausibility of the resulting architecture also matters: if a computational solution was selected by evolution, it probably deserves attention (no pun intended). Of course, this selection may just be the result of biophysical (e.g. metabolic) constraints that are not relevant to computer vision. But conversely, it could well be that brain-inspired solutions represent a true functional optimum towards machine intelligence; and that the dominant strategy in the field, of iteratively optimizing deep learning architectures with SOTA accuracy as the sole objective, could be driving us towards a local minimum in the space of functional architectures. Here, we look to the brain for inspiration on alternative attention architectures for computer vision.

\vspace{-2mm}
\subsubsection{Visual attention in the brain -}
How is attention implemented in the brain, and how does it differ from current deep learning models in computer vision? 

The first thing to note is that deep convolutional networks are, to a first approximation, fairly representative of the computations taking place in the first feedforward sweep of information through the hierarchy of visual brain regions~\cite{fukushima1988neocognitron,riesenhuber1999hierarchical}. As neural information propagates from the retina through the thalamus, the primary ``striate'' visual cortex, and subsequent ``extra-striate'' visual areas, towards temporal cortex regions where object recognition and categorization take place~\cite{logothetis1995shape}, the pattern of synaptic connections between neurons undergoes a systematic increase of receptive field sizes, spatial invariance, and complexity of the neuron's optimal features (from small oriented edges in V1, to full objects or scene classes in infero-temporal cortex). This pattern is compatible with what one would expect from a series of convolutional kernels in deep learning models. 

This apparent match between deep convolutional networks and the feed-forward sweep of neural activity in the brain~\cite{kriegeskorte2015,yamins2016} does not mean that attention plays no role in vision---only that attention typically comes into play \emph{after} this initial feed-forward sweep. In this sense, attention in the brain is thus very different from the way that it has been recently inserted into deep convolutional networks~\cite{wang2017,bello2019} or implemented by vision transformers~\cite{ramachandran2019,zhao2020,dosovitskiy2020image,touvron2020training}, as a direct component of the main feed-forward pass.

The brain comprises a separate and unified attention network (the so-called ``fronto-parietal'' network) that receives sensory inputs from the various brain regions, and on the basis of this information determines where and how to pay attention~\cite{corbetta2002control,szczepanski2013functional,itti2001computational}. Subsequently, attention signals from the fronto-parietal network are sent back to the visual stream to modulate neural activations according to attentional priorities~\cite{desimone1995neural,schafer2011selective,hamker2005reentry}. This is of course a very coarse description that leaves aside important nuances, but in short, attention in the brain is computed outside of the visual cortical hierarchy, based on its initial feedforward activation, and modulating it in an iterative fashion at subsequent steps. 
Could a similar architecture also benefit deep convolutional networks?

\vspace{-2mm}
\section{Proposed architecture}
\vspace{-2mm}
Here we present a series of simple experiments to begin addressing this question. Our base architecture is a deep convolutional network pre-trained for image classification, which we augment with a separate attention system, and an iterative attention modulation mechanism implementing a form of ``routing by agreement''. Attention priority is computed as a matching (or agreement) score between keys and queries, as in Transformer architectures. Here, for simplicity, the queries are pooled across the entire network, effectively implementing a form of Global Attention Agreement (hence the name: GAttANet). While this greatly simplifies the computational demands of the attention network (compared to a full self-attention strategy), this simple first implementation proved sufficient to improve classification accuracy across a variety of backbones and datasets. 

\begin{figure}[h!]
\includegraphics[width=\textwidth]{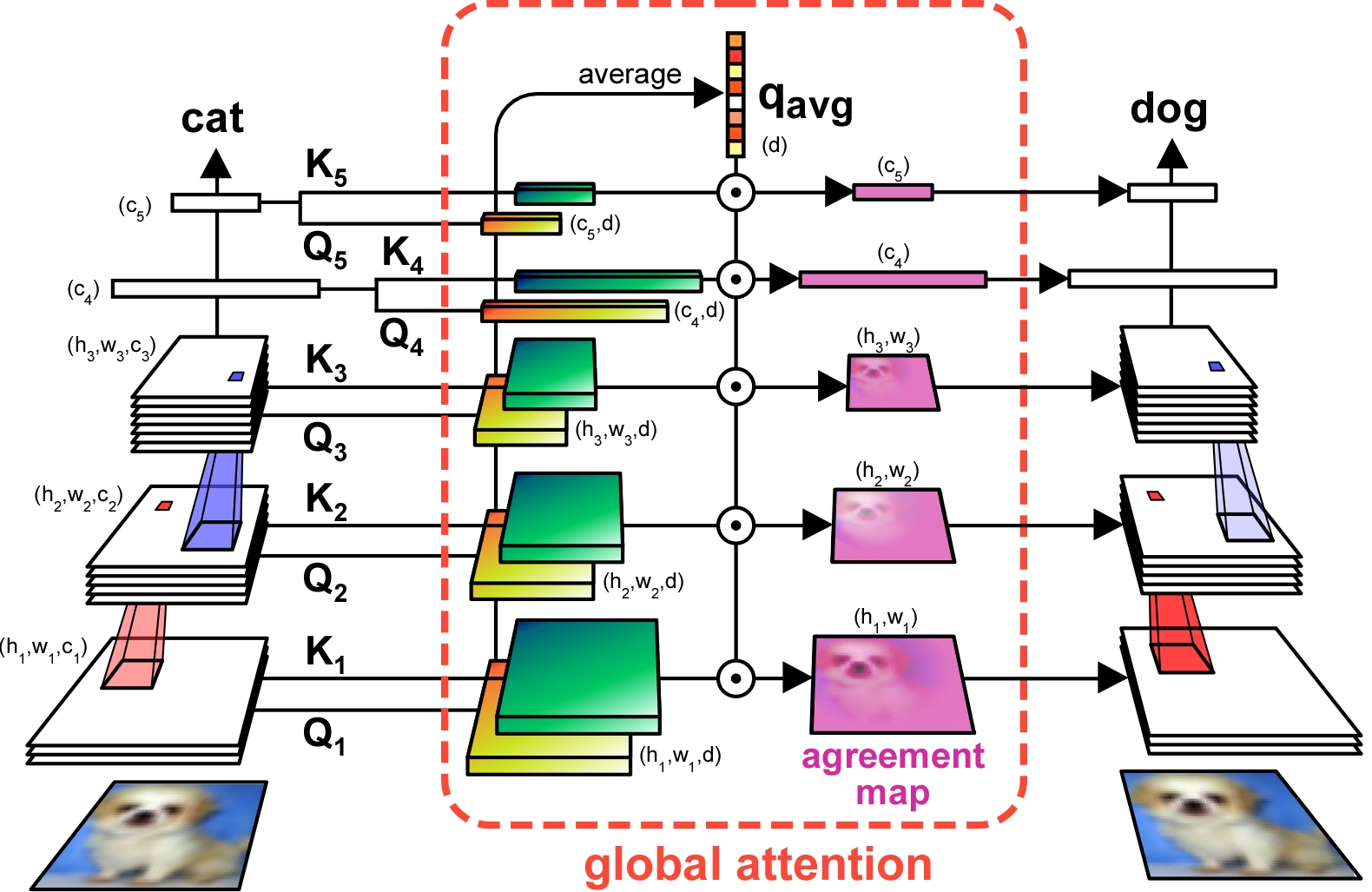}
\caption{Proposed GAttANet architecture. A standard convolutional network (here our ``toy model'' with 3 convolutional layers and 2 dense layers) can be augmented with the global attention agreement system. Each layer's activations are multiplied by Q and K matrices to produce corresponding Query and Key maps. Queries are averaged across all layers and spatial positions, resulting in a unique $q_{avg}$ global query vector. The dot product of this global query with each layer's key determines the layers global attention agreement map, that directly modulates the layer's activations on the next time step. Features that were more or less compatible with the rest of the network are up- or down-regulated (respectively; compare red and blue pixels in the initial vs. final states), and the network's classification can be improved. } 
\vspace{-2mm}
\label{architecture}
\end{figure}

\vspace{-2mm}
\subsubsection{Convolutional network backbones -}
Our first backbone was a ``toy model'' with three convolutional layers followed by two dense (fully-connected) layers, the latter of which served as the classification layer. This was meant as a computationally inexpensive architecture to explore our proposed augmentation with global attention, its individual components and its functional properties. Specifically, we used (3x3) convolutional kernels at each stage, ReLU activation functions, followed by (2x2) max pooling operations to decrease spatial resolution, and 0.2 drop-out as regularization. The input RGB image corresponded to 3 input channels, and the three subsequent convolutional layers comprised respectively 32, 64 and 128 channels. After flattening the output of the last convolutional layer, it was projected onto a dense layer with 256 units (with ReLU activation), then onto a final dense layer for classification, ending with a softmax operation (the number of classes in the dataset, 10 for CIFAR10 and 100 for CIFAR100, dictating the corresponding number of units). The resulting networks, counting around 620,000 to 640,000 parameters (depending on the dataset), were pretrained on CIFAR10 or CIFAR100. We used data augmentation (0-20\degree rotation, 0-20\% shift in width and height, and random horizontal flips), with a batch size of 128, the Adam optimizer (with default parameters) and early-stopping (patience=50 epochs) until convergence, which typically took less than 500 epochs. The resulting baseline networks reached an accuracy of 83.28\% on CIFAR10 and 52.54\% on CIFAR100 (see Table~\ref{tab1}). These baseline architectures were then augmented with our global attention agreement mechanism, as described in the next section.

To determine the usefulness of our proposed scheme in more general situations, we also explored standard modern convolutional architectures as backbones, namely ResNet18 and ResNet50, pretrained on ImageNet-1k. These networks comprised respectively 11.7M and 25.6M parameters. The baseline top-1 accuracy was 68.43\% for ResNet18 and 74.94\% for ResNet50 (see Table~\ref{tab1}).

\vspace{-2mm}
\subsubsection{Global Attention Agreement -}
The general architecture of our proposed Global Attention Agreement system (GAttANet) is illustrated in Figure \ref{architecture}. Just like in the brain, attention is envisioned here as a separate system from the convolutional visual backbone, that receives information from it (in the form of attention keys and queries) and influences its processing in return (based on the global attention agreement score at each location in the network). 

More specifically, the activation in each layer $i$ is turned into a pair of (Key, Query) attention maps ($k^i$,$q^i$), via learned linear projection matrices $K_i$ and $Q_i$. This is done slightly differently for convolutional and dense layers. For a convolutional layer $Conv^i$ of spatial dimensions $(h_i,w_i)$ and $c_i$ channels, the attention matrices are $K_i,Q_i \in \mathbb{R}^{c_i \times d}$ where $d$ is the chosen attention dimension (in our experiments, we varied $d$ between 4 and 64). The attention projection is given by the dot product:
\begin{equation}
\begin{aligned}
    k^i(x,y,m) &= \sum_{c=1}^{c_i}(Conv^i(x,y,c)*K_i(c,m)) \\
    q^i(x,y,m) &= \sum_{c=1}^{c_i}(Conv^i(x,y,c)*Q_i(c,m))
\end{aligned}
\end{equation}
where $(x,y) \in [1,w_i]\times[1,h_i]$ is a unit's spatial position, and $m \in [1,d]$.

For a dense layer $Dense^j$ of $c_j$ units,  the learned attention matrices are $K_j,Q_j \in \mathbb{R}^{c_j \times d}$, and the attention projection is given by the scalar product:
\begin{equation}
\begin{aligned}
    k^j(c,m) &= Dense^j(c)*K_j(c,m) \\
    q^j(c,m) &= Dense^j(c)*Q_j(c,m)
\end{aligned}
\end{equation}
where $c \in [1,c_j]$ is a unit's index, and $m \in [1,d]$.

All queries inside the network (across all layers and spatial positions) are then collected and averaged into a single global query $q_{avg}$, as follows:
\begin{equation}
\begin{aligned}
    q_{avg}(m) &= \frac{1}{n_{c}+n_{d}}\left(\sum_{i \in \mathbb{C}}\sum_{x=1}^{w_i}\sum_{y=1}^{h_i}\frac{q^i(x,y,m)}{w_i*h_i}  + \sum_{j \in \mathbb{D}}\sum_{c=1}^{c_j}\frac{q^j(c,m)}{c_j}\right)
\end{aligned}
\end{equation}
where $m \in [1,d]$, and $\mathbb{C}$ and $\mathbb{D}$ are the index sets of convolutional layers and dense layers, respectively, of cardinals $n_c$ and $n_d$.

Next, the $q_{avg}$ global query is compared against each key across the entire network by means of a simple dot product, resulting in the ``Global Attention Agreement'' score:
\begin{equation}
\begin{aligned}
    gatta^i(x,y) &= k^i(x,y,.) \cdot q_{avg} \\
    gatta^j(c) &= k^j(c,.) \cdot q_{avg}
\end{aligned}
\end{equation}
for convolutional and dense layers, respectively.

On the next pass through the network, each unit's computation is modulated (via multiplicative scaling) according to the global attention agreement score assigned to it, i.e.:
\begin{equation}
\begin{aligned}
    Conv^i(x,y,c) &:= Conv^i(x,y,c)*(1+\alpha_i*gatta^i(x,y))  \\
    Dense^j(c) &:= Dense^j(c)*(1+\alpha_j*gatta^j(c))
\end{aligned}
\end{equation}
where $\alpha_i$ is a learned parameter controlling the strength of attentional modulation for each layer $i$.
 
\vspace{-2mm}
\paragraph{ResNet models} are augmented in a slightly different way compared to the toy model. To limit computational demands, only a subset of layers are connected to the global attention system. Specifically, four convolutional layers are chosen to span the model's hierarchy (typically the output layer of a ResNet block), plus two dense layers: the average-pooling layer and the final classification layer (pre-softmax). These chosen layers, and only these, convey keys and queries to the global attention system, and receive attentional modulation in return, as described in Equations (1-5).

\vspace{-2mm}
\subsubsection{Training details -}
We trained the global attention system on the original datasets (CIFAR10 or CIFAR100 for the toy model, ImageNet-1k for the ResNet models), with the pretrained weights of the convolutional backbone entirely frozen. Thus, the only trained parameters were the key/query matrices $K_i$ and $Q_i$ across the entire network, and the attentional modulation factor $\alpha_i$ for each layer. The number of trained parameters was very small in comparison to the number of weights in the backbone models (see Table \ref{tab1}): with $d=16$, there were about 20,000 parameters to train for the toy model, 100,000 for the ResNet18 backbone and 200,000 for the ResNet50 (compared to about 600,000, 12M and 25M weights, respectively). During training we applied 0.25 drop-out to keys and queries for regularization in all models; for the toy models, we also used 0.00001 $L_2$ regularization; for ResNet models, we applied batch-normalization for keys and queries, and layer normalization for the $gatta$ attention scores. We used a batch size of 128 for the toy models and 8 for the ResNets; the Adam optimizer with learning rate set to 0.001 (the default) for the toy models and 0.0003 for the ResNets; and early-stopping (patience=500 epochs for toy models and 1 epoch for ResNets) until convergence, which typically took less than 1000 epochs for toy models and less than 10 epochs for ResNets.

\begin{table}[h!]
\setlength{\tabcolsep}{4pt}
\caption{\textbf{Accuracy across models and datasets.} The base parameters are pretrained and fixed (corresponding accuracy listed under `base acc.'), and we only train the additional attentional parameters (`att. params'). The final accuracy is listed under `acc. (ours)', in bold for the best accuracy over a given backbone/dataset combination.}\label{tab1}
\vspace{-2mm}
\begin{tabular}[c]{m{1.6cm} m{2cm} m{1.4cm} m{1.2cm} m{1.1cm} m{1.4cm} m{1.4cm}}
\toprule
\vspace{-4mm}
\textbf{backbone} & \textbf{dataset} & \textbf{base params.} & \textbf{base acc.} & \textbf{att. dim.} & \textbf{att. params.} & \textbf{acc. (ours)} \\
\toprule
toy model & CIFAR10 & 620.4K & 83.28\% & \begin{tabular}{@{}l@{}}$d=16$ \end{tabular} & 15.8K & \textbf{85.34\%}\\
\midrule
toy model & CIFAR100 & 643.5K & 52.54\% & \begin{tabular}{@{}l@{}}$d=16$ \\ $d=32$ \\ $d=64$ \end{tabular} & \begin{tabular}{@{}l@{}} 18.7K \\ 37.4K \\ 74.9K \end{tabular} & \begin{tabular}{@{}l@{}}55.54\% \\ 55.86\% \\ \textbf{56.03\%} \end{tabular} \\
\midrule
ResNet18 & ImageNet-1k & 11.70M & 68.43\% & \begin{tabular}{@{}l@{}} $d=8$ \\ $d=16$ \\ $d=32$ \end{tabular} & \begin{tabular}{@{}l@{}} 68.9K \\ 101.4K \\ 166.5K \end{tabular}& \begin{tabular}{@{}l@{}} 68.72\% \\ 68.83\% \\ \textbf{68.84}\% \end{tabular}\\
\midrule
ResNet50 & ImageNet-1k & 25.64M & 74.94\% & \begin{tabular}{@{}l@{}} $d=4$ \\ $d=8$ \\ $d=16$ \\ $d=32$ \\ $d=64$ \end{tabular} & \begin{tabular}{@{}l@{}} 78.7K \\ 118.0K \\ 196.7K \\ 353.9K \\ 668.4K \end{tabular} & \begin{tabular}{@{}l@{}} \textbf{75.23\%} \\ 75.21\% \\ 75.18\% \\ 75.20\% \\ 75.18\% \end{tabular}\\
\bottomrule
\end{tabular}
\vspace{-6mm}
\end{table}

\vspace{-2mm}
\section{Results}
\vspace{-2mm}
\subsubsection{Accuracy -} Table \ref{tab1} summarizes test set accuracy for the different models and datasets. The proposed GAttANet architecture yields accuracy improvements over the toy models on the order of 2\% for CIFAR10 and up to 3.5\% for CIFAR100. Given the relatively small increase in parameters, this improvement is noticeable (informal tests with feedforward toy models using comparable parameter numbers, obtained by augmenting the number of convolution channels, did not yield any significant improvement). 

For ResNets, performance improvements were robust but more modest: about 0.3-0.4\% on ImageNet top-1 accuracy. Obviously, it may be more difficult to optimize a ResNet---a pretty solid model already---compared to our simple toy model. Still, these improvements are not negligible, especially considering the small number of additional parameters. If we use as a reference the slope of the function relating parameters to accuracy between a ResNet18 and a ResNet50, the measured accuracy improvement for a standard ResNet architecture would have required 0.9M additional parameters (8.5 times more than our proposed architecture with $d$=16). Similarly. using the ResNet50-ResNet101 slope as a reference, it would take a standard ResNet architecture with 3.8M additional parameters to match our augmented version of ResNet50 (which is 48 times more than our proposed architecture with $d$=4). Therefore, our approach appears viable not just in simple scenarios, but also in state-of-the-art models. 

Nonetheless, to limit computational demands, the following explorations of the global attention system were performed with the (more flexible) toy models.

\begin{figure}[h!]
\vspace{-4mm}
 \begin{center}
\includegraphics[width=0.8\textwidth]{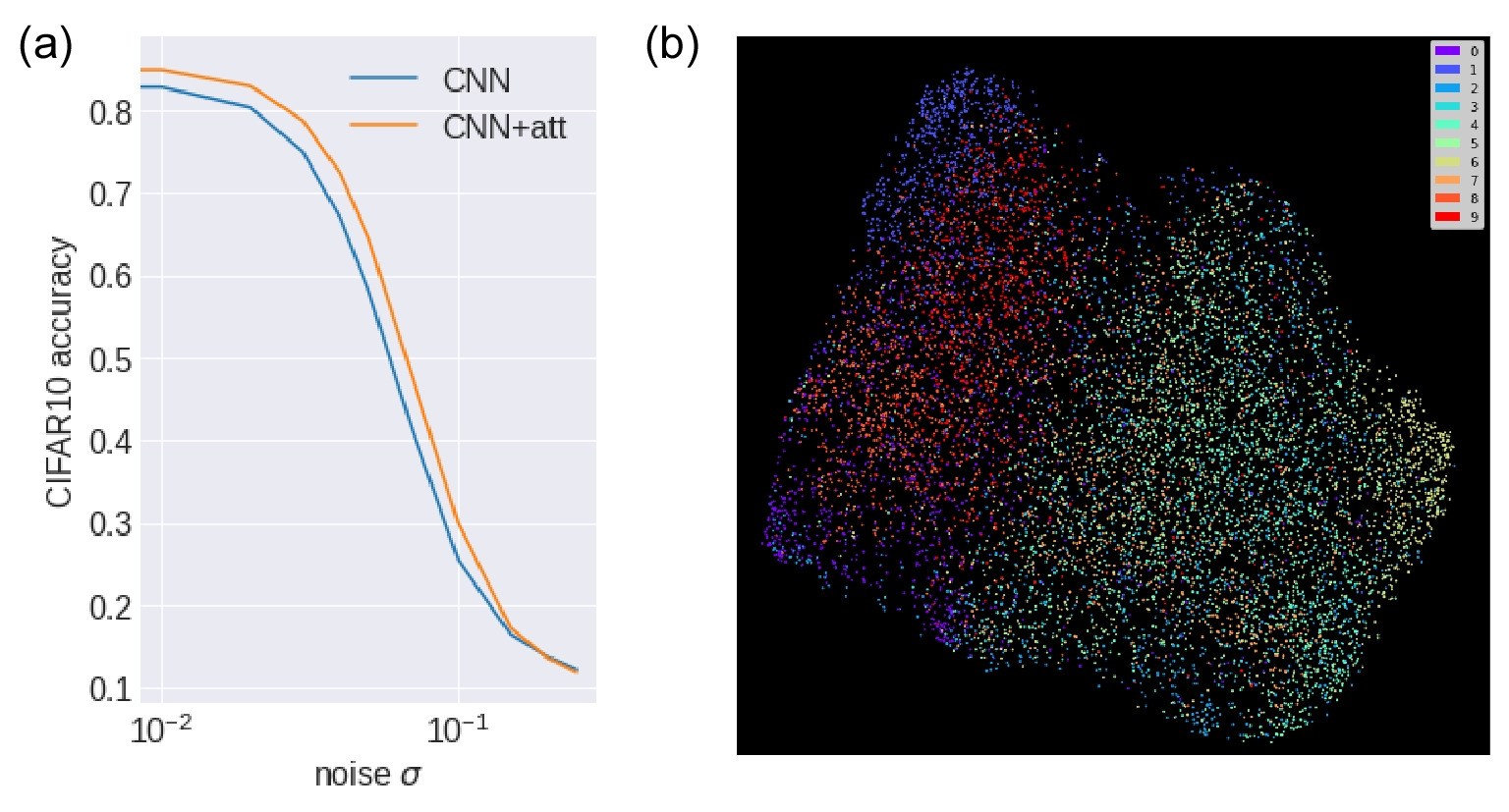}
\end{center}
\vspace*{-6mm}
\caption{Results on CIFAR10. (a) Comparison between the baseline model and the attention-augmented model ($d$=16). Accuracy is plotted as a function of the amount of Gaussian noise (log-scale) added to each image. (b) 2-D UMAP visualization of the learned attention space of the global average query $q_{avg}$. Points are colored by image category (in alphabetical order). Animal classes (bird, cat, deer, dog, frog, horse) correspond to labels 2-7 and project in a separate region compared to vehicle classes (airplane, automobile, ship, truck) with labels 0-1 and 8-9.} \label{cifar10}
\vspace{-4mm}
\end{figure}

\vspace{-4mm}
\subsubsection{Noise robustness -}
To assess the generalization abilities of our proposed strategy with respect to out-of-distribution examples, we exposed the trained models to various levels of additive Gaussian noise on input images. For both CIFAR10 (Figure \ref{cifar10}a) and CIFAR100 (Figure \ref{cifar100}a-b), the performance improvements from our global attention strategy on clean images remained consistently visible across several levels of noise, even increasing for moderate noise (up to 8 percentage point improvements on CIFAR100 with noise $\sigma$=0.03 to 0.04), and only vanishing when the baseline model (without attention) approached chance level (noise $\sigma$=0.1 to 0.25).

\vspace{-2mm}
\subsubsection{Properties of the Global Attention query space -}
Figure \ref{cifar10}b shows a 2-D embedding of the learned 16-D space of the global attention query $q_{avg}$ across images of the CIFAR10 dataset. It is noteworthy that the 10 classes appear to be separated, in particular along a main direction reflecting the `animals vs. vehicles' distinction. Although such a separation may already exist in the representation layers of the convolutional backbone, the fact that it is also visible in the global attention system indicates that it has learned (through the key and query matrices) a meaningful representation of image properties. We believe that this could make the system useful not only in a bottom-up attention scenario as here (where the input fully determines the attentional modulation), but also in a top-down attention scenario, where the model's behavior could be controlled by the user in a class-specific way, e.g. when there is a strong prior for a given class (like `airplane'), or for a given semantic property (like `vehicle vs. animal'). We plan to explore this avenue in follow-up work.

\begin{figure}[h!]
\vspace{-2mm}
\begin{center}
\includegraphics[trim=0 0 0 20, clip,width=\textwidth]{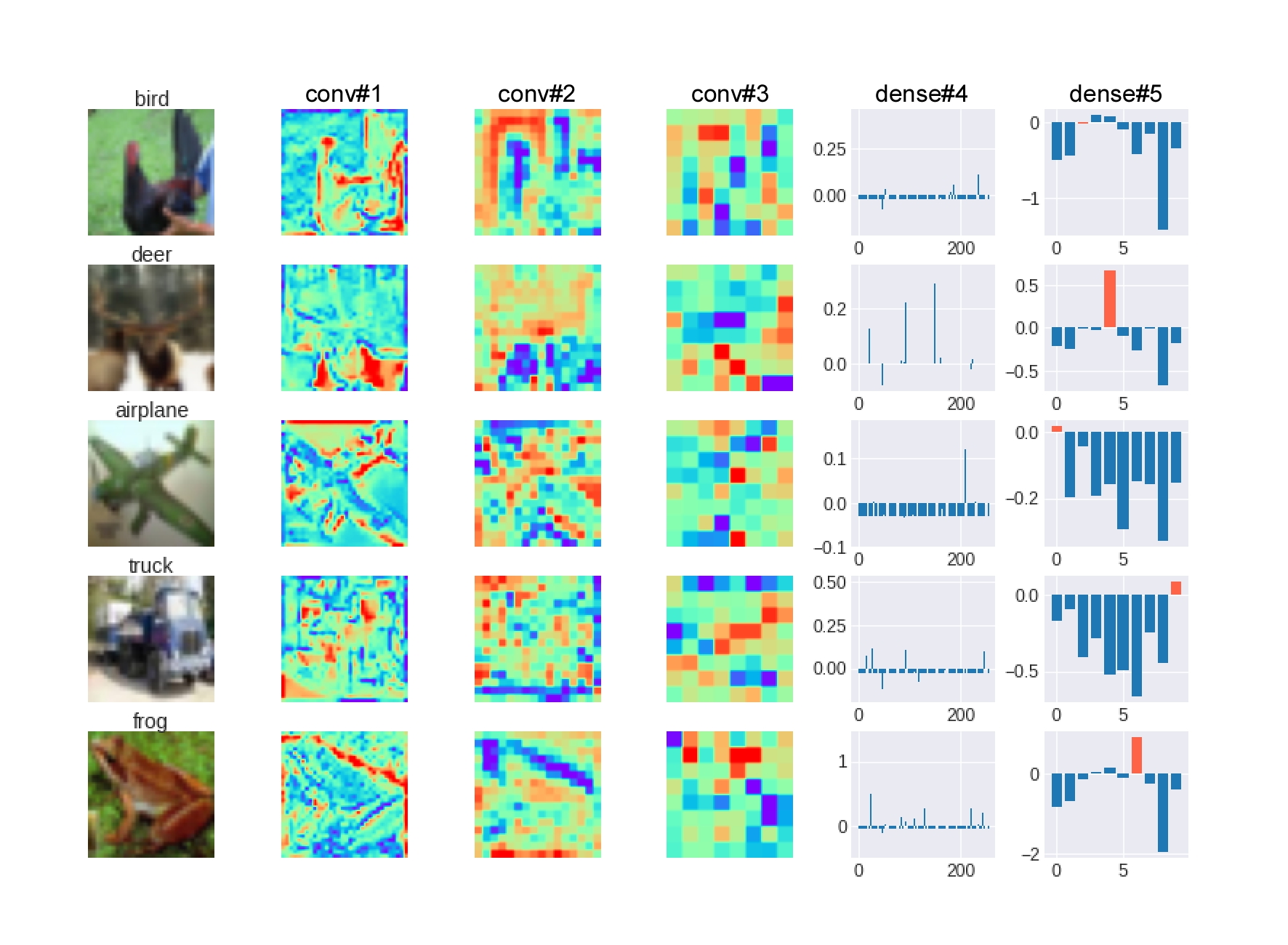}
\end{center}
\vspace*{-10mm}
\caption{Visualization of attention agreement scores. For the three convolutional layers, the score is represented as a spatial map, for the two dense layers as a bar plot (neurons along the x-axis). The last dense layer reflects the agreement of the global query $q_{avg}$ with the key from each possible class of CIFAR10. The correct class, highlighted in red, is often the one with highest agreement. } \label{visualization}
\vspace{-2mm}
\end{figure}

\vspace{-2mm}
\subsubsection{Visualization of global attention agreement maps -}
In addition to the learned space of the global attention query $q_{avg}$, it may be helpful to visualize global attention agreement maps for each of the model's layers. The resulting map visualizations in Figure \ref{visualization} correspond to the $gatta$ maps defined in Eq (4), and schematically illustrated in purple color in Figure~\ref{architecture}. We see that in the final layer, the logit corresponding to the target class is typically among the units with the highest global attention agreement score (red bar in the right-most column). This indicates that on the next feed-forward pass through the convolutional network, this unit's activity will be increased by attention. Similarly, many spatial locations in the 2D agreement maps will be specifically enhanced (red colors) or decreased (blue colors) by attention.

\vspace{-2mm}
\subsubsection{Effect of attention dimension $d$ -} 
\begin{figure}[h!]
\vspace{-0mm}
 \begin{center}
\includegraphics[width=\textwidth]{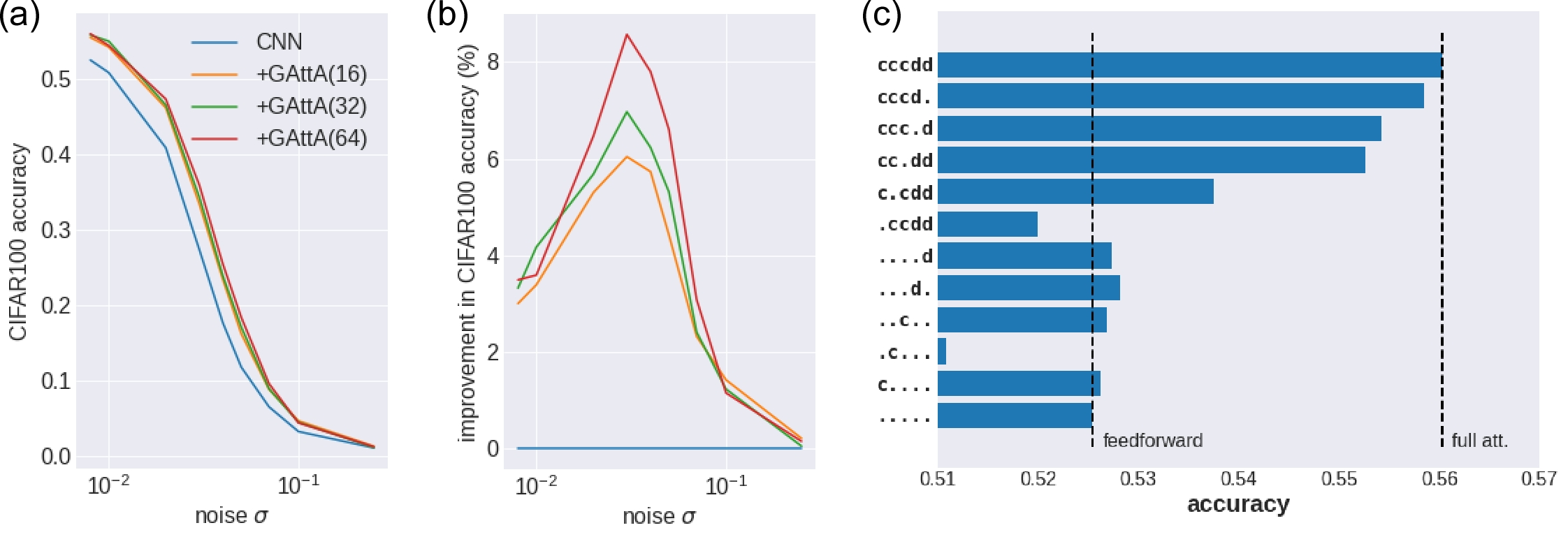}
\end{center}
\vspace*{-6mm}
\caption{Results on CIFAR100. (a) Comparison between the baseline CNN and our attention-augmented models (with different dimensions $d$ for the attention key/queries). Accuracy is plotted as a function of the amount of Gaussian noise (log-scale) added to each image. (b) Improvement in accuracy of the attention-augmented models relative to the baseline CNN (c) Lesion studies of the trained model ($d=16$). After training, we ran the model with certain layers receiving no attention modulation (layers indicated by a `.' on the left; for example, `c.cdd' indicates that the second convolutional layer did not receive attention, while `...d.' indicates that only the first dense layer received attention). No single layer was sufficient to yield performance improvements in isolation; but some layers impaired the model more than others when they were lesioned (e.g., compare `.ccdd' to `cccd.'). } \label{cifar100}
\vspace{-4mm}
\end{figure}

Table \ref{tab1} as well as Figure \ref{cifar100}b provide a mixed interpretation for the effects of increasing the dimension $d$ of the key/query attention space. On the one hand, going from $d$=16 to $d$=64 proved beneficial for the toy model on CIFAR100, and even more so at moderate levels of added Gaussian noise (Figure \ref{cifar100}b). On the other hand, for ResNet models on ImageNet there was not much effect of increasing $d$, at least in the range explored. Our global attention strategy was only marginally better for a ResNet18 with $d$=16 or 32 compared to $d$=8, and was equally (or more) beneficial for a ResNet50 with $d$=4 as with $d$=64. This may be because ResNet backbones are already close to optimal, and there is little room for improvement. 

\vspace{-2mm}
\subsubsection{Lesion experiments -} 
Finally, we asked if attentional modulation of specific convolutional or dense layers in our toy model was critical to the observed attentional improvements. Figure \ref{cifar100}c reports CIFAR100 accuracy of the baseline toy model (denoted by `.....' in the figure, i.e. no layer receiving attentional modulation); the full attention-augmented version (`cccdd', all 3 convolutional layers and 2 dense layers receiving attentional modulation); and several ablations of the latter. First, when a single layer's modulation was ablated (marked by a `.' in the figure), effects on the model performance varied from inexistant (`cccd.') to catastrophic (`.ccdd'). However, when only a single layer was modulated at a time (all other modulations ablated), no performance improvements were visible. This indicates that, while some layers may be more important than others, the global attention agreement strategy requires pooling signals across the entire model's hierarchy.

\vspace{-2mm}
\section{Discussion}
\vspace{-2mm}
We described an attention architecture with a global key/query matching system that pools queries across the entire hierarchy of layers in a convolutional network, and in return modulates each layer's activations based on their global attention agreement score. This  proposal is similar to---and in large part also inspired by---the many ``vision transformer'' architectures in the recent literature~\cite{ramachandran2019,zhao2020,dosovitskiy2020image,touvron2020training}, especially those that employ transformer modules \emph{in addition to} (not \emph{instead of}) convolutional layers~\cite{wang2017,bello2019}. However, a fundamental difference is that our transformer/attention system is entirely separate from the convolutional backbone---just like the frontoparietal attention system in the brain is separate from the visual regions that it modulates~\cite{corbetta2002control,szczepanski2013functional,desimone1995neural,schafer2011selective,hamker2005reentry}. 

Our system improved classification accuracy compared to each convolutional backbone across multiple datasets, at a minimal cost in terms of additional parameters. Augmenting the feedforward convolutional backbones with a similar parameter budget (i.e. with additional convolutional channels or ResNet blocks) would not produce significant performance improvements (we explicitly tested this for the toy models). It is possible that stand-alone transformer vision models would be more parameter-efficient~\cite{dosovitskiy2020image,touvron2020training}, but we view this as an orthogonal question: even if a vision transformer could match or surpass our attention model, it remains a biologically implausible architecture.

Does our global attention system relate to the ``Global Workspace'' framework for cognition and awareness, advocated by several authors in cognitive science~\cite{baars2005global,shanahan2006cognitive}, neuroscience~\cite{mashour2020}, and more recently also in machine learning~\cite{bengio2017,vanrullen2020deep}? As initially proposed, the Global Workspace is a shared multimodal representation used to collect relevant information across multiple independent neural systems, and to broadcast its contents to the rest of the brain. Our model is a unimodal attention system and as such, does not really fit this description. On the other hand, we note that our proposed architecture is highly similar in implementation to the ``Shared Global Workspace'' recently described by Goyal and colleagues~\cite{goyal2021coordination}. In their model, each stage of a hierarchical visual Transformer architecture sends and receives information from a separate ``workspace'' module via a key-query attention mechanism, essentially similar to our global attention system (except for our use of convolutional layers instead of transformer modules). Thus, while we view our proposed attention model as independent from the Global Workspace theory, it may serve as a building block for a future large-scale Global Workspace system.

Several possibilities come to mind for improving our system in the future. For instance, rather than relying on a pretrained network with frozen weights, the convolutional backbone could be trained (or at least fine-tuned) jointly with the global attention system. In addition, could we further increase performance by iterating the global attention agreement mechanism across multiple time steps? For the present model trained on a single iteration, our explorations revealed that this was actually detrimental. A model \emph{trained} for two or more iterations could still outperform ours, but our initial attempts in this direction encountered difficulties in terms of computational demands and numerical stability, so we leave this question open for future work.

Could we use a full self-attention mechanism rather than relying on our global query $q_{avg}$? That is, compute the entire pairwise map of attention agreement scores between all network locations? Theoretically yes, though this would also require a separate scheme for combining activation values across distinct layers having potentially different channel numbers. This could be achieved, for example, by relying on \emph{value} matrices $V_i$, as in standard Transformers. In practice, however, this could prove prohibitively costly, as the computational cost of full self-attention grows with $O(N^2)$, instead of $O(N)$ for our pooled global attention (where $N$ is the number of spatial locations in the network). Yet this is definitely one avenue to explore in the future.

Finally, could we also benefit from using \emph{multi-head} attention---or an equivalent strategy allowing attention to simultaneously query multiple input properties? The potential functional advantage in the context of global attention agreement could be an ability for the network to simultaneously ``agree'' on multiple objects or interpretations. This may be particularly helpful in ambiguous situations, or for images with numerous targets. Such a scheme, unfortunately, appears incompatible with our current system where the final output is a scalar attention modulation score for each location. In ongoing work, we are exploring the possibility of employing complex-valued units, whose phase angles can be controlled by pairwise mutual attention agreement, to achieve a similar purpose. Our hope is that convolutional networks augmented in this way may develop a form of ``binding-by-synchrony'', whereby clusters of complex phase values delimit distinct objects in the scene, as observed in several neuroscience experiments~\cite{singer2001consciousness,fries2015rhythms}.
\vspace{-2mm}

\bibliographystyle{splncs04}
\bibliography{refs}
%
%
%
%
%
\end{document}